# Causal Mosaic: Cause-Effect Inference via Nonlinear ICA and Ensemble Method

**Pengzhou (Abel) Wu**
The Graduate University for Advanced Studies

**Kenji Fukumizu**
The Institute of Statistical Mathematics

## Abstract

We address the problem of distinguishing cause from effect in bivariate setting. Based on recent developments in nonlinear independent component analysis (ICA), we train nonparametrically general nonlinear causal models that allow non-additive noise. Further, we build an ensemble framework, namely Causal Mosaic, which models a causal pair by a mixture of nonlinear models. We compare this method with other recent methods on artificial and real world benchmark datasets, and our method shows state-of-the-art performance.

## 1 INTRODUCTION

Causal discovery (Spirtes and Zhang, 2016; Peters et al., 2017) is a fundamental problem which attracts increasing attention recently. The golden standard of causal discovery is randomized controlled experiments, but they often encounter ethical and practical issues. Thus, causal discovery from pure observational data provides an indispensable way to understand our nature. Traditionally, causal discovery algorithms learn the causal structure in the form of a directed acyclic graphical (DAG) model, by searching in the space of possible DAGs (Drton and Maathuis, 2017). Constraint-based search methods (e.g. FCI (Spirtes et al., 2000)) use conditional independence tests to determine the causal structure. Score-based search methods, such as GES (Chickering, 2002), typically search for a graph that optimizes a penalized likelihood score. However, the above methods are unable to fully determine causal directions.

In recent years, a line of research emerges that is particularly motivated to solve the problem of distinguishing cause from effect in bivariate case, i.e. cause-effect inference. All these methods exploit cause-effect asymmetry to identify causal direction (Mooij et al., 2016). One major approach is to restrict causal mechanism to a certain class of "functional causal models" (FCMs) (Hyvärinen and Zhang, 2016), and the causal direction between $C$ and $E$ is identifiable if $p(E|C)$ can be fitted by this class, while the opposite direction, $p(C|E)$, cannot. Typical FCMs are LiNGAM (Shimizu et al., 2006), ANM (Hoyer et al., 2009), PNL (Zhang and Hyvärinen, 2009) and ANM-MM (Hu et al., 2018). And all of them assume additive noise. Many other methods loosely exploit the idea that the process generating cause distribution $p(C)$ is in some way "independent" to the causal mechanism generating conditional distribution $p(E|C)$. For example, IGCI (Janzing et al., 2012) uses orthogonality in information space to express the independence between the two distributions. KCDC (Mitrovic et al., 2018) is based on the invariance of Kolmogorov complexity of conditional distribution. RECI (Blöbaum et al., 2018) extends IGCI to the setting with small noise, and proceeds by comparing the regression errors in both possible directions.

This work studies cause-effect inference, with focus on the following limitations in existing methods. First, FCMs put too strong restrictions to the functional form of causal mechanism. Second, other works tend to propose simple "principles" that actually reflect the authors' own intuition on causality. Thus, most of existing methods failed to achieve high accuracy on real world data. Third, there are a few methods (e.g. KCDC, CGNN (Goudet et al., 2018)) that use more flexible models and achieve better performance, but without theoretical justifications.

We address the problems respectively as follows. First, we train nonlinear causal models on cause-effect pairs with (maybe partial) direction information, based on a recent nonlinear ICA method implemented by neural network, without strong restriction on the functional relationship among the variables or the noise structure. Second, the fact that each of the many approaches to causality works to some limited extent suggests us to take a "mosaic" view: causal systems are diverse and heterogeneous, so we should not fit all the different systems at once; instead, study at a time a small number of causal systems that share common aspects, and then build a whole picture. Specifically, we build an ensemble of nonlinear models, which amounts to a Causal Mosaic: a causal pair's mechanism is treated as a mixture of similar mechanisms. It is analogous to constructing a large piece of mosaic from tesserae, which are small blocks of material used in creating a mosaic. Finally, we provide theoretical results on the conditions under which our method will work.

The main contributions of this paper are : 1) two novel cause-effect inference rules with identifiability proofs, 2) an ensemble framework that works for real world datasets with only limited labeled pairs, 3) a neural network structure designed for causal-effect inference, and 4) state-of-the-art performance on real-world benchmark dataset.

**Related work** RCC (Lopez-Paz et al., 2015) and its follow-up NCC (Lopez-Paz et al., 2017) also use training data, but they require large numbers of labeled pairs and thus rely on synthetic pairs for training. There is work which takes related viewpoints: KCDC uses majority voting, the simplest ensemble method; ANM-MM treats mechanism as a mixture. NonSENS (Monti et al., 2019) also employs the same nonlinear ICA method as ours, but needs samples of a casual system available over different environments, which requires interventions or even experiments. We should note that all the above methods neither take a mosaic view explicitly nor use ensemble method as a main building block.

## 2 PRELIMINARIES

### 2.1 Intuition

As mentioned, we encounter a large diversity of causal relationships in nature. And causality might only be studied and learned piecemeal. Our idea is to extract the common mechanism shared by a small number of causal systems. We should note that, systems that seem to have different mechanisms can actually share the same mechanism. When all we have at hand is observational data, the sample, it would be true that two systems sharing the same mechanism, but by looking at the samples, they seem very different, to the extent that we would be tempted to model them by different functional forms. As an example, we give some pairs we used in experiment in Figure 1.

In the following subsections, we first formally introduce our problem setting, then show its connection to nonlinear ICA, and finally review the nonlinear ICA method which we exploit to learn shared mechanism.

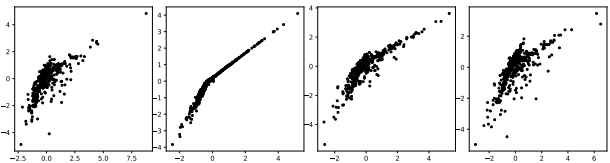

Figure 1: Artificial causal pairs sharing same mechanism. The pairs have significant diversity though still show some regularity. Please refer to Section 5 for details.

### 2.2 Notation and Problem Setting

Generally, causal inference problems can be formalized by Structural Causal Models (SCMs) (Pearl, 2009), also known as Structural Equation Models (SEMs) (Bollen, 1989). Let $\mathcal{G} = (\mathcal{V}, \mathcal{E})$ denote a causal DAG, where $\mathcal{V}$ is the vertex set and $\mathcal{E}$ is the edge set. Then, the SCM of observed variables $\mathbf{X} = (X_v)_{v \in \mathcal{V}}$ and *independent* hidden variables $\mathbf{E} = (E_v)_{v \in \mathcal{V}}$ is given by the set of equations [1]:

$$X_v = f_v(X_{pa_\mathcal{G}(v)}, E_v), v \in \mathcal{V} \qquad (1)$$

$f_v$ represents the *causal mechanism* between effect $X_v$ and its direct causes (parents in the graph) $X_{pa_\mathcal{G}(v)}$. And $E_v$ models exogenous (external) influences on $X_v$ and is often treated as an unobserved noise.

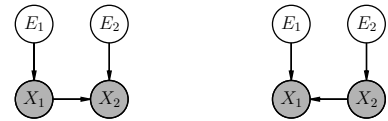

Figure 2: Causal graphs of bivariate SCMs

In this work, we focus on bivariate cases, where there are only two possibilities: either $X_1$ or $X_2$ is the direct cause of the other, as shown in Figure 2. Their SCMs are the following (2) for $X_1 \to X_2$, and (3) for $X_2 \to X_1$. In cause-effect inference, our goal is to distinguish between these two possibilities, that is, tell cause from effect.

$$X_1 = f_1(E_1), \qquad X_2 = f_2(X_1, E_2) \qquad (2)$$
$$X_1 = f_1(X_2, E_1), \qquad X_2 = f_2(E_2) \qquad (3)$$

### 2.3 Nonlinear ICA and Causal Discovery

A straightforward definition of the generative model for nonlinear ICA is that independent hidden variables $\mathbf{Z} = (Z_1, ..., Z_n)$ are mixed by a differentiable and invertible nonlinear function $\mathbf{f}$, and produce observed variables $\mathbf{X} = (X_1, ..., X_n) = \mathbf{f}(\mathbf{Z})$. The goal is to

---

[1] As typical definition of SCM, we rule out *feedback* loops (two-way causal influences) and *confounders* (hidden common causes) here.

recover the independent components $Z_i$ and the unmixing function $\mathbf{g} = \mathbf{f}^{-1}$, only using observations of $\mathbf{X}$. The following definition formally states the connection between SCM and nonlinear ICA:

**Definition 1.** An SCM (1) is **analyzable** if there exists a differentiable and invertible function $\mathbf{f} : \mathbf{R}^n \to \mathbf{R}^n$, such that $\mathbf{X} = \mathbf{f}(\mathbf{E})$.

Obviously, an analyzable SCM is a special case of nonlinear ICA's generative model, with particular structure between the variables. For example, in bivariate SCM (2), let $f_3(E_1, E_2) = f_2(f_1(E_1), E_2)$ and $\mathbf{f} = (f_1, f_3)$, the SCM can be written as $(X_1, X_2) = \mathbf{f}(E_1, E_2)$. Now if $\mathbf{f}$ is differentiable and invertible on $\mathbf{R}^2$, the SCM is analyzable.

For analyzable SCM, if we can solve the corresponding nonlinear ICA problem, we obtain the hidden variables $\mathbf{E} = \mathbf{g}(\mathbf{X})$. In bivariate case, given $E_1$ and $E_2$, under causal Markov and faithfulness assumptions (Spirtes and Zhang, 2016), we can conclude:

$$\begin{aligned} X_1 \to X_2 \text{ if } X_1 \perp\!\!\!\perp E_2, \\ X_2 \to X_1 \text{ if } X_2 \perp\!\!\!\perp E_1 \end{aligned} \quad (4)$$

This criteria was exploited by many classical methods, e.g. LiNGAM and ANM, and can be easily understood as the independence of noise and cause.

### 2.4 Learning Shared Mechanism by TCL

Lately, Time-Contrastive Learning (TCL) (Hyvärinen and Morioka, 2016) provided the first general identifiability result for nonlinear ICA. The method depends on learning the different distributions of time series through time, and hence the name. After artificially dividing time series into segments, it trains a classification task to tell which segment each sample point belong to. As indicated in Hyvärinen et al. (2019), the segment index could be treated as an *auxiliary variable* $\mathbf{u}$, which only needs to satisfy that hidden components $\mathbf{Z}$ are independent of each other given $\mathbf{u}$.

With the intuition that different causal pairs in real world can share the same mechanism, we can derive a method for learning the shared mechanism by TCL. We just need to feed TCL with pairs sharing mechanism, and replace segment index with pair index as auxiliary variable. Here, we restate the theory under our own setting:

**Theorem 1** (Hyvärinen and Morioka (2016)). *Assume the following:*

*A1. We observe causal pairs $\mathcal{X}(P) := \{\mathbf{X}_p\}_{p=1}^P$ which satisfy the same analyzable SCM $\mathbf{X}_p = \mathbf{f}(\mathbf{E}_p)$, and the hidden variables $E_{i,p}, i = 1, 2$ are of exponential family distribution $p_{E_{i,p}}(e) = \exp[T_i(e)\eta_i(p) - A(\eta_i(p))]$ where $T_i(e)$ is the sufficient statistic.*

*A2. The matrix $\mathbf{L}$, with elements $[\mathbf{L}]_{p,i} = \eta_i(p) - \eta_i(1))$, $p = 1, ..., P$, $i = 1, 2$, has full column rank 2.*

*A3. We train a feature extractor $\mathbf{h} : \mathbf{R}^2 \to \mathbf{R}^2$ with universal approximation capability, followed by a final softmax layer to classify all sample points of the pairs, with pair index used as class label.*

*Then, in the limit of infinite data, for each $p$, $\mathbf{T}(\mathbf{E}_p) := (T_1(E_{1,p}), T_2(E_{2,p}))^T = \mathbf{A}\mathbf{h}(\mathbf{X}_p; \boldsymbol{\theta}) + \mathbf{b}$ where $\mathbf{A}, \mathbf{b}$ are unknown constants, and $\mathbf{A}$ is invertible.*

In practice, a multilayer perceptron (MLP) is used as the feature extractor. The theorem implies that the identification (recovery) of $\mathbf{T}(\mathbf{E}_p)$ can be achieved by first performing TCL, and then linear ICA on $\mathbf{h}(\mathbf{X}_p)$. Denoting the composition of $\mathbf{h}$ and linear ICA as $\mathbf{hICA}$, we have $\mathbf{T}(\mathbf{E}_p) = \mathbf{hICA}(\mathbf{X}_p)$. In this sense, we say that $\mathbf{h}$ is successfully learned and the nonlinear ICA of $\mathbf{X}_p$ is *realized* by $\mathbf{hICA}$. Here we learn the shared mechanism $\mathbf{f}$ (or precisely its inverse) as part of $\mathbf{h}$, along with $\mathbf{T}$.

While we can recover only the sufficient statistics $T_i(E_{i,p})$, not $E_{i,p}$, they are sufficient for building a method for cause-effect inference; $T_i(E_{i,p})$ generally has the same independence relationships with other variables as $E_{i,p}$. In practice, under the assumption that there exist direct causal effects, we can just compare values of an independence measure, as we will detail in Section 3.

Unlike the time contrast exploited in the original TCL, the contrast here is among the pairs. But, by convention, we will still use the word *"TCL"* when referring to the method trained on causal pairs, which *not* necessarily satisfy A1 and A2 of Theorem 1. By a slight abuse of terminology, the produced $\mathbf{h}$, which may *not* be successfully learned, is also called TCL in this paper.

## 3 THEORETICAL RESULTS

### 3.1 Separation of Training and Testing

It should be clear from Section 2 that we want to learn casual mechanism via TCL. However, readers might notice that, to successfully learn TCL, we at least need to know that the pairs indeed share causal mechanism! To address the above dilemma, our idea is to learn causal mechanism from some training pairs that we have good causal knowledge (e.g. we might know their SCMs and causal directions), and then predict the causal directions for unseen pairs. The following corollary of Theorem 1 makes this separation possible:

**Corollary 1** (**Transferability** of TCL). *Assume:*

*A1. Pairs $\mathcal{X}^{tr}(P)$ satisfy A1 and A2 of Theorem 1.*

*A2. A pair $\mathbf{X}^{te}$ satisfy A1 of Theorem 1, with the same $\mathbf{f}$ and $\mathbf{T}$ as $\mathcal{X}^{tr}(P)$, but different parameter $\eta_i$.*

*A3. Let $\mathcal{R}_X$ denote the support of a random variable $X$. We have $\mathcal{R}_{E_i^{te}} \subseteq \cup_{p=1}^{P} \mathcal{R}_{E_{i,p}^{tr}}, i = 1, 2$.*

*A4. We learn a feature extractor $\mathbf{h}$ on $\mathcal{X}^{tr}(P)$ as in A3 of Theorem 1 and have $\mathbf{T}(\mathbf{E}_p^{tr}) = \mathbf{A}\mathbf{h}(\mathbf{X}_p^{tr}) + \mathbf{b}$.*

*Then, we have $\mathbf{T}(\mathbf{E}^{te}) = \mathbf{A}\mathbf{h}(\mathbf{X}^{te}) + \mathbf{b} = \mathbf{hICA}(\mathbf{X}^{te})$.*

Intuitively, after we successfully learned TCL $\mathbf{h}$, we can re-use it to analyze other unseen pairs that have the same SCM and sufficient statistics as the training pairs. We should note that, as in transfer learning, training and testing pairs do *not* have the same distribution, and hence the name of this corollary. From now on, we will also refer to the learning of TCL and analysis of new pairs on it as training and testing, respectively.

### 3.2 Inference Methods and Identifiability

With $\mathbf{T}(\mathbf{E}^{te})$ recovered, we can find ways to infer a causal direction for $\mathbf{X}^{te}$. To find the asymmetry between the two possible causal directions, we use the fact that, when testing, if we *flip* input direction to **hICA** and try nonlinear ICA for each, there will be one and only one trial that is realized by the **hICA**.

A remaining issue is that, to apply Theorem 1 and in turn Corollary 1, we need to know the directions of training pairs. (This is implied by $\forall p\, (\mathbf{X}_p = \mathbf{f}(\mathbf{E}_p))$ in A1 of Theorem 1.) More precisely, $\{\mathbf{X}_p\}_{p=1}^{P}$ must be *aligned*, as in the following definition.

**Definition 2.** Causal pairs $\{\mathbf{X}_p\}_{p=1}^{P}$ are **aligned** if $\forall p\, (X_{1,p} \to X_{2,p})$ or $\forall p\, (X_{2,p} \to X_{1,p})$.

Below, we will give two inference rules with their identifiability results, based respectively on specific conditions that enable the alignment of pairs. Before detailing the two rules, we present a general procedure (Algorithm 1) as the common basis. In the following, $\alpha_0 = (1, 2)$ and $\alpha_1 = (2, 1)$ denotes the two permutations on $\{1, 2\}$, and $\sigma$ an unknown permutation taking value on $\{\alpha_0, \alpha_1\}$. The pairs with *unknown* directions are denoted by $\boldsymbol{\sigma}(\mathcal{X}(P)) := \{X_{\sigma_p(1),p}, X_{\sigma_p(2),p}\}_{p=1}^{P}$.

---

**Algorithm 1:** Inferring causal direction

**input** : $\boldsymbol{\sigma}(\mathcal{X}^{tr}(P))$, $\boldsymbol{\sigma}(\mathbf{X}^{te})$, $Direction^{tr}$, `align`, `inferule`
**output:** $Cause^{te}$
1 Align training set, exploiting $Direction^{tr}$:
   $\mathcal{X}^{al}(P) = \texttt{align}(\boldsymbol{\sigma}(\mathcal{X}^{tr}(P)), Direction^{tr})$
2 Learn TCL $\mathbf{h}$ on $\mathcal{X}^{al}(P)$
3 **foreach** $\alpha = \alpha_0, \alpha_1$ **do**
4 $\quad (C_1, C_2)_\alpha^T = \mathbf{hICA}(X_{\alpha(1)}^{te}, X_{\alpha(2)}^{te})$
5 Run inference rule:
   $Cause^{te} = \texttt{inferule}(\mathbf{C}_{\alpha_0}, \mathbf{C}_{\alpha_1}, \boldsymbol{\sigma}(\mathbf{X}^{te}))$

---

In the first inference rule, it is assumed that we know the causal direction for each of the training pairs so that they can be trivially aligned. For a test pair, a realized (successful) nonlinear ICA among the two trials should output independent components, and this tells us the direction from cause to effect. This leads to the following theorem:

**Theorem 2** (**Identifiability** by independence of hidden components)**.** *In Algorithm 1, let:*

$Direction^{tr} = \{c_p\}_p^{p=P}$ *where* $c_p \in \{1, 2\}$ *is the cause index:* $X_{c_p,p}^{tr} \to X_{3-c_p,p}^{tr}$,

$\texttt{align} = \{X_{c_p,p}^{tr}, X_{3-c_p,p}^{tr}\}_p^{p=P}$,

$\texttt{inferule} = \alpha^*(1), \alpha^* = \underset{\alpha \in \{\alpha_0, \alpha_1\}}{\arg\max}\, \texttt{dindep}(\mathbf{C}_\alpha)$ *where* `dindep` *measures degree of independence.*

*And assume:*

*A1. Causal Markov assumption and causal faithfulness assumption hold for data generating SCMs and analysis procedure except[2] for a realized nonlinear ICA.*

*A2. $\mathcal{X}^{tr}(P)$ and $\mathbf{X}^{te}$ satisfy A1–A3 of Corollary 1.*

*Then, the `inferule` defined above (`inferule1` afterwards) identifies the true cause variable.*

In the second inference rule, we examine the independence of the pair $\{T_j(E_j^{te}), X_i^{te}\}$, as in the relation (4). Note, however, that as described in Monti et al. (2019), the outputs of a realized nonlinear ICA are equivalent to hidden variables only up to a permutation, i.e. $\mathbf{T}(\mathbf{E}^{te}) = (C_{\alpha(1)}, C_{\alpha(2)})^T$, with $\alpha$ unknown. This requires us to evaluate the degree of independence for four pairs at each trial, as stated in the following theorem:

**Theorem 3** (**Identifiability** by independence of noise and cause)**.** *In Algorithm 1, let:*

$Direction^{tr} = \{\sigma_p\}_p^{p=P}$ *where $\sigma_p$ is the permutation for pair $\mathbf{X}_p$,*

$\texttt{align} = \{X_{\sigma_p^{-1}(\sigma_p(1)),p}^{tr}, X_{\sigma_p^{-1}(\sigma_p(2)),p}^{tr}\}_{p=1}^{P} = \mathcal{X}^{tr}(P)$,

$\texttt{inferule} = i^*, (i^*, ., .) = \underset{i,j,\alpha}{\arg\max}\, \texttt{dindep}(X_i^{te}, C_{j,\alpha})$.

*And assume the same as Theorem 2.*

*Then, the `inferule` defined above (`inferule2` afterwards) identifies true cause variable.*

`inferule2` determines the realized trial and identifies causal directions, with only aligned training set and *without* the causal directions of training pairs (the $Direction^{tr}$ in Theorem 2). Since we can use the causal directions to recover an aligned training set, so in Theorem 2, letting `inferule = inferule2`, the true causal index can also be identified. However, as we will see in the experiments, `inferule1` will outperform `inferule2` if the former is applicable in practice.

---
[2]See Supplementary Materials on this.

Finally, we will employ distance correlation (dCor) (Székely et al., 2007) as our main choice of `dindep`[3].

### 3.3 Structural MLP

We discuss an MLP structure to improve TCL's performance on bivariate analyzable SCMs. We first study the form of the inverse SCM, since this is what the MLP should learn.

**Proposition 1** (Inverse of bivariate analyzable SCM). *For any analyzable SCM as shown in* (2), *denote the whole system* $\mathbf{X} = \mathbf{f}(\mathbf{E})$, *if the Jacobian matrix of* $\mathbf{f}$ *is invertible, then* $f_1$ *is invertible.*

Denote $g_1 = f_1^{-1}$, then $E_1 = g_1(X_1)$. And we have $E_2 = g_2(X_1, X_2)$ in general. This implies the inverse SCM has the graph as shown in Figure 3 (left):

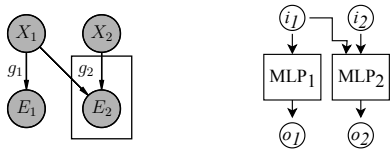

Figure 3: Inverse bivariate analyzable SCM (left) and the indicated MLP structure (right).

Building an MLP for TCL with this asymmetric structure will help TCL learn the inverse SCM. This can be easily implemented as shown in Figure 3 (right): we build an MLP with one output node for $g_1$ and $g_2$ respectively, and then concatenate the outputs together. While one might think that we need to make MLP1 invertible since $g_1$ is invertible, we should *not* impose it; the sufficient statistics $\mathbf{T}$ are also learned as part of MLP, and they are in general non-invertible.

Finally, there are two caveats. First, the structural MLP works only when there is a direct causal effect, as required by SCM (2). Second, since node $i_1$ corresponds to the cause, we need to input the cause variable to $i_1$ for training the asymmetric MLP properly. This requires knowledge on the causal directions of training pairs, and thus, we can only apply it with `inferule1`.

## 4 ASSEMBLING CAUSAL MOSAIC

In the following, we will refer to training pairs that satisfy A1 and A2 of Theorem 1 (same SCM and exponential family) as *tessera pairs*, because they form the small portion of causal pairs that can be easily modeled together, and thus a small block of the whole mosaic. Also, we will refer to a TCL learned on tessera pairs as a *tessera*.

We have so far assumed that we have tessera pairs, under the ideal situation that we have well-studied systems. However, for many real world applications, it is unlikely that most training pairs amount to tessera pairs. Our idea for handling real world problems is to train many TCLs on random selections of pairs, and then choose from these TCLs the ones that are tesserae, which may not be perfect. We develop an ensemble method to effectively exploit imperfect tesserae.

In this section, Let $S$ be the set of all labeled causal pairs we have at hand, and $c_s$ be the true cause index for $s \in S$.

### 4.1 Preparing Materials

As in Algorithm 2, by training a large number ($N$) of TCLs on randomly chosen pairs, we hope some of these TCLs amount to tesserae. To ensure TCL is trained properly on each set of pairs, we train MLP $M$ times with different hyperparameters (See experiment for details).

---
**Algorithm 2:** Random training of TCLs

**input** : $S, M, N$
**output**: $\{(\mathbf{h}_n, T_n)\}_{n=1}^N$

1 **foreach** *n in 1,...,N* **do**
2     Randomly choose training pairs $T_n \subset S$
3     Split the sample points of each training pair by half, and build training set $Tr$ and testing set $Te$
4     **foreach** *m in 1,...,M* **do**
5       Randomly choose a set of hyperparameters and train TCL on $Tr$
6       Evaluate classification accuracy ($Cacc_m$) for pair index on $Te$.
7     Use the trained TCL with the highest $Cacc_m$ for this set of training pair, denote it $\mathbf{h}_n$

---

### 4.2 Making Tesserae

Because our goal is to infer causal directions, we choose TCLs that perform well on this task. First, we can use each TCL to infer the causal directions of its own training pairs (Algorithm 3, line 2,3)[4], and choose TCLs that produce accuracy higher than a threshold $ThreT$. If the training pairs consist of exact tessera pairs, by the identifiability theorems, their directions should be correctly inferred. Second, for each TCL, we also input unseen validation pairs and infer their directions, and we choose TCLs that produce accuracy higher than $ThreV$. The good training accuracy indicates the suc-cess of training and TCL indeed learned to infer causal directions. The good validation accuracy shows that the learning generalizes to unseen pairs.

---
[3]See Supplementary Material for details.

[4]Can be simplified, similarly to Algorithm 5, line 2–4.

To efficiently use $S$ for training and validation, and still be able to test on all the pairs in $S$, we use the idea of leave-one-out cross validation (LOOCV). That is, each pair $l$ not used in training a TCL is left out once when validating that TCL (line 5,6).

---

**Algorithm 3:** Selecting TCLs

**input** : $S$, $ThreT$, $ThreV$, $\{(\mathbf{h}_n, T_n)\}_{n=1}^N$
**output:** $\{TSR_s : s \in S\}$

1 **foreach** $n$ in $1,...,N$ **do**
   // Training accuracy $Tacc_n$ for $\mathbf{h}_n$ on $T_n$
2    **foreach** $t$ in $T_n$ **do**
3       Use $\mathbf{hICA}_n$, run line 3–5 of Algorithm 1 on $t$, get inferred direction $\hat{c}_t$
4    $Tacc_n = |\{t : \hat{c}_t = c_t\}|/|T_n|$
   // LOOCV
5    **foreach** $l$ in $S \setminus T_n$ **do**
6       As line 2–4, get validation accuracy for $\mathbf{h}_n$ on $(S \setminus T_n) \setminus \{l\}$, denote it as $Vacc_n(l)$
   // Select TCLs by accuracy thresholds
7 **foreach** $s$ in $S$ **do**
8    Initialize tessera index set for $s$: $TSR_s = \varnothing$
9    **foreach** $n$ in $1,...,N$ **do**
10       **if** $s \notin T_n$ and $Tacc_n > ThreT$ and $Vacc_n(s) > ThreV$ **then**
11          Add $n$ to $TSR_s$

---

As we can see, every pair in $S$ is not used as training pair or validating pair for its tessera (line 10,11). On the other hand, in training ($T_n$) and validation (($S \setminus T_n) \setminus \{l\}$), every trained TCL exploits all the pairs except the left out one $l$.

### 4.3 From Tesserae to Causal Mosaic

Even if we have selected TCLs as in Algorithm 3, it is very possible that the chosen tesserae would not be perfect, e.g., the mechanisms of training pairs are not exactly the same. Consequently, learned commonality in an imperfect tessera does not fully describe the mechanisms of the pairs, but only partially.

Accordingly, we employ an ensemble method for making effective use of each imperfect tessera, and construct a whole piece of mosaic, in the same way as we will obtain a strong classifier from weaker ones by ensemble methods. Simply put, for each testing pair, ensemble method will take the results from tesserae, and produce a final, weighted average.

The weighting method is the core of an ensemble method, and we want to weight each tessera for *each* testing pair. Thus, we introduce two levels of weighting, both based on $\texttt{dindep}(\mathbf{hICA}(s))$ for pair $s$. First, a weight that measures how well the training pairs of TCL fit together, it is the average $\texttt{dindep}(\mathbf{hICA}(.))$ for the training pairs (Algorithm 4, line 3). Second, we weight by the $\texttt{dindep}(\mathbf{hICA}(.))$ for a testing pair on a TCL. As in Algorithm 1, in theory only realized nonlinear ICA outputs independent components, so we weight by the larger $\texttt{dindep}$ of the two trials (line 4–7). We multiply the two weights as the final pair-specified weight.

---

**Algorithm 4:** Ensemble method

**input** : $S$, $\{TSR_s : s \in S\}$, $\{(\mathbf{h}_n, T_n)\}_{n=1}^N$
**output:** $\{Direction_s : s \in S\}$

1 **foreach** $s$ in $S$ **do**
2    **foreach** $n$ in $TSR_s$ **do**
3       $w_n = \sum_{t \in T_n}(\texttt{dindep}(\mathbf{hICA}_n(t)))/|T_n|$
4       **foreach** $i = 0, 1$ **do**
5          $\mathbf{C}_{\alpha_i} = \mathbf{hICA}_n(\alpha_i(s))$
6          $w_{ns,i+1} = \texttt{dindep}(\mathbf{C}_{\alpha_i})$
7       $w_{ns} = \max(w_{ns,1}, w_{ns,2})$
8       $\hat{c}_s = \texttt{inferule}(\mathbf{C}_{\alpha_0}, \mathbf{C}_{\alpha_1}, s)$
9       $Direction_{ns} = 1$ if $\hat{c}_s = 1$, $-1$ if $\hat{c}_s = 2$
10    Calculate weighted prediction
    $Score_s = \sum_{n \in TSR_s} w_n w_{ns} Direction_{ns}$
11    $Direction_s = \begin{cases} X_1 \to X_2 & Score_s > 0 \\ X_2 \to X_1 & Score_s < 0 \\ ? & Score_s = 0 \end{cases}$

---

## 5 EXPERIMENTS

### 5.1 Artificial Data

We compare NonSENS to variations of our method with different inference rules, independence measures, and MLP types on artificial data. To see the comparisons with other recent methods on similar artificial data, we refer readers to Monti et al. (2019).

**Multi-environment setting** This is the setting under which NonSENS works. Mathematically, our tessera pairs $\{\mathbf{X}_p^{tr}\}$ are equivalent to the samples $\mathcal{X}^{en} := \{\mathbf{X}_p^{en}\}$ of a *same* causal system under $P$ different "environments" in their interpretation. That is, they define different environments by different parameter $\boldsymbol{\eta}$ of hidden variables, and $\forall p(\mathbf{X}_p^{en} = \mathbf{f}(\mathbf{E}_p^{en}))$ is by definition satisfied. Moreover, there is no separate testing pairs here. Our goal is to distinguish between two possibilities, $\forall p\,(X_{1,p}^{en} \to X_{2,p}^{en})$ or $\forall p\,(X_{2,p}^{en} \to X_{1,p}^{en})$, for $\mathcal{X}^{en}$ themselves (note the pairs (environments) are *aligned*), rather than $2^P$ possibilities for individual pairs $\mathcal{X}(P)$.

Our Algorithm 1 can reduce to this setting, as shown in Algorithm 5. Both training and testing pairs are $\mathcal{X}^{en}$ themselves. Note that $Direction^{tr}$, $\texttt{align}$ and the input permutation (Algorithm 1, line 3,4) are not needed, since $\mathcal{X}^{en}$ is already aligned. We apply a simplified version of $\texttt{inferule2}$ to infer direction for each environment without input permutation, but still need to deal with the output permutation.

Finally, we use majority voting to combine the results of all environments and give the final decision, and this is an important difference between our method and NonSENS under this setting. NonSENS treats the samples of environments as coming from a mixture, runs dindep on pooled sample and output, and gives $c^{en} = i^*, (i^*, j^*) = \mathrm{argmax}_{i,j}\, \mathtt{dindep}(\{X_{i,p}^{en}\}, \{C_{j,p}\})$[5]. In practice, as we will see, majority voting often outperforms NonSENS since it uses information from each environment and thus is more robust.

---

**Algorithm 5:** Algorithm 1 on multi-environment setting

**input** : $\mathcal{X}^{en}$
**output**: $c^{en}$

1 Learn TCL $\mathbf{h}$ on $\mathcal{X}^{en}$
2 $\mathcal{C} = \mathbf{hICA}(\mathcal{X}^{en})$
3 **foreach** $\mathbf{X}_p^{en}\ in\ \mathcal{X}^{en},\ \mathbf{C}_p\ in\ \mathcal{C}$ **do**
4 $\quad c_p = i^*, (i^*, j^*) = \underset{i,j}{\mathrm{argmax}}\, \mathtt{dindep}(X_{i,p}^{en}, C_{j,p})$
  // Majority voting
5 $c^{en} = \underset{i}{\mathrm{argmax}}\, |\{c_p : c_p = i\}|$

---

**Multi-pair setting** If we know the *directions* of training pairs, we separate training and testing, and both Theorem 2 (`inferule1`) and Theorem 3 (`inferule2`) can apply. Here, we infer the direction for each individual testing pair. NonSENS cannot apply here, so we compare different variations of our method. We name this multi-pair setting, to contrast the multi-environment setting, although the main difference is the direction information of training pairs (our method *can* also infer for each environment as in Algorithm 5, line 3,4).

**Data generation** We generate data in the same ways as in Hyvärinen and Morioka (2016) and Monti et al. (2019). We use a 5-layer MLP as the mixing function, with leaky ReLU activation and 2 units in each layer to ensure invertibility. To simulate the independent relationships of a direct causal graph, we use lower-triangle weight matrix for each layer of the MLP. We use Laplace distribution for both hidden components, and their variance parameters differ though different pairs. We can easily simulate multi-environment setting by aligning all the artificial pairs and then perform nonlinear ICA.

We generate 100 different mixing functions, and for each mixing function, we generate same number of training pairs. Under multi-pair setting, we use the same testing pairs as multi-environment setting, and we keep the same number of training and testing pairs. To observe how the number of training pairs affect results, we try 5 different number ranging from 10 to 50.

**Hyperparameters** For the MLP of TCL, we use the same number of layers as data-generating MLP, having the same number of units (4 or 40 in the experiments) for all hidden layers with the maxout activation. For the asymmetric MLP (Figure 3, right), we use the same width for both sub-MLPs, and keep the sum of the widths the same as fully-connected MLP. Note that the asymmetric MLP has much less parameters than the fully-connected one, since the sub-MLPs are disconnected. The two output units have the absolute value function as activation. To make fair comparisons, for both our method and NonSENS, we keep all the hyperparameters the same, including the training parameters (e.g. learning rate and batch size) and the threshold of independent tests (e.g. alpha value).

**Assuming direct causal effect** Our method and NonSENS[6] formally requires direct causal effects exist between pairs, and this is our main experiment setting. Please see Supplimentary Material for the experiment without this assumption.

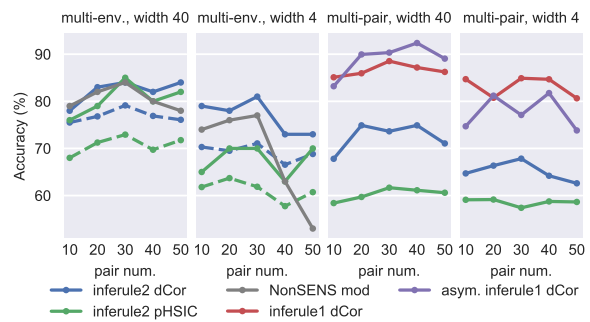

Figure 4: Performance assuming direct causal effect. In the legend, "dCor/HSIC" indicates the independence measure we use, and "asym." means asymmetric MLP in TCL.

As shown in Figure 4, in multi-environment setting, our method largely outperforms NonSENS expect for large pair (environment) number. A possible reason is that, as mentioned, NonSENS runs dindep on pooled sample, and HSIC works better with more sample points.

In multi-pair setting, `inferule1` is applicable and performs much better than `inferule2`. The main reason is that the independence between two output components is much easier to realize than the independence between estimated noise and observed cause. And this is in turn because of the direct dependence between observed variables and outputs (see Figure 1 in Supplementary Material). Note that Theorem 2 re-quired known causal directions of training pairs, and thus cannot be used in multi-environment setting.

---

[5]Originally, NonSENS uses independence tests with a threshold. We write it here using dindep for easy comparison, because we will use this modified rule for NonSENS in experiment.

[6]We cannot reproduce the likelihood ratio based NonSENS proposed for this setting. Instead, we use a slightly modified version of NonSENS originally proposed for may-not-direct-causal setting, see footnote 5.

Moreover, when the MLP width is 40, `inferule1` achieves near optimal results when applied with asymmetry MLP. This is also the best result we have obtained with artificial data. While the asymmetry MLP with width 4 performs worse than the fully-connected one, this is due to the limited fitting capacity (see Supplementary Material for details).

To confirm the transferability of TCL, we also try inferring directions for individual pairs without voting under multi-environment setting (Figure 4, dashed lines). The results from the two settings are similar, meaning the transferability. The slight drop of performance under multi-pair setting should come from the two input trials needed.

When inferring by Theorem 3, we try both dCor and the p-value of HSIC (Gretton et al., 2005) as `dindep`. dCor constantly outperforms HSIC (See Supplementary Material for details).

### 5.2 Real World Dataset

Tuebingen cause-effect pairs (TCEP) dataset (Mooij et al., 2016)[7] is a commonly used benchmark for cause-effect inference tasks. Causal Mosaic can be suitably applied here because of the very diverse scenarios of the pairs. Each pair is assigned a weight in order to account for the possible correlation with other pairs that are selected from the same multivariate scenario. Currently the dataset contains 108 real-world cause-effect pairs with true causal directions labeled by human experts. We exclude 6 multivariate pairs in our evaluation.

We use Theorem 2 with asymmetric MLP since it already shows much better results on artificial data. Unlike on artificial data with Laplace hidden variables, we use maxout activation for the output layer. When implementing Algorithm 4 line 10, we use a simplified version $Score_s = \sum_{n \in TSR_s}(w_{ns,1} - w_{ns,2})$, since this works the best. See Supplementary Material for details.

We train TCL on 300 ($N$) sets of randomly picked pairs, which are of size ranging from 4 to 32. The TCLs are trained 10 ($M$) times on each pair set. For selecting TCLs, we randomly search 100 pairs of accuracy thresholds ($ThreT, ThreV$) in $[65\%, 75\%]^2$ and rule out too large thresholds that give 0 or only 1 tessera for more than 10 TCEP pairs. During training, the following hyperparameters are randomly chosen from uniform distributions: depth and width of MLP, learning rate, decay factor, max. steps, momentum, and batch size. Among them, the depth of MLP larger than 10 might lead to divergence in training, but the other parameters seem to have little impact if we do not use some extreme values.

Table 1: Accuracy (%) on TCEP. "A/B" means with/without applying pairs weight.

| ANM | IGCI | RECI | NCC | OURS |
|---|---|---|---|---|
| 52.5/52.0 | 60.4/60.8 | 70.5/62.8 | 51.8/56.9 | $\mathbf{81.5}_{\pm 4.1}/\mathbf{83.3}_{\pm 5.2}$ |

We compare our method to ANM, IGCI, RECI and NCC[8], the results are shown in Table 1. We report the *median* and std-error of accuracies of our method calculated on all the 83 pairs of thresholds. And this already shows state-of-the-art performance. The best result on all thresholds is 86.3% and might overfit TCEP dataset. For NCC, we infer each pair by training the method on rest of the pairs. The accuracy is much worse than that reported in Lopez-Paz et al. (2017), the most possible reason is that NCC requires much more training data (10,000 artificial pairs in the original paper). The performance of ANM is worse than reported in Mooij et al. (2016), possibly because of the different implementation of independence test.

## 6 CONCLUSION

In this work, we proposed a highly flexible cause-effect inference method that learns a mixture of general nonlinear causal models, with proof of identifiability. We exploited TCL to extract the common mechanism shared by different causal pairs, and transferred the causal knowledge to unseen pairs. More specifically, our method learns how to distinguish cause from effect, from some training pairs, and predicts the causal direction on testing pairs. We gave two inference rules with identifiability proofs and an ensemble framework that works on real world cause-effect pairs with limited labeled causal directions. We compared our method to other state-of-the-art methods on artificial and real world benchmark dataset, and it showed favorable results.

Hence, we justified the "mosaic" perspective of causal discovery, which proposes to learn causality piecemeal, and then build a whole picture by the pieces. Here, shared mechanism learned by TCL forms a tessera of the whole causal mosaic, and many tesserae are learned and further combined into a whole picture by ensemble method. We believe this new perspective would promote other novel methods for bivariate and also more general causal discovery problems.

---

[7] We use the latest version on December 20, 2017.

[8] We use the implementations from CDT package (Kalainathan and Goudet, 2019)

# 7 Proofs

**Corollary 1**

*Proof.* From A4, and substitute $\mathbf{X}_{\text{p}}^{\text{tr}} = \mathbf{f}(\mathbf{E}_{\text{p}}^{\text{tr}})$, we have $\mathbf{T}(\mathbf{E}_{\text{p}}^{\text{tr}})^{\text{T}} = \mathbf{A}\mathbf{h}(\mathbf{f}(\mathbf{E}_{\text{p}}^{\text{tr}})) + \mathbf{b}$.

From A3, we know each $\mathbf{X}^{te}$'s support is contained in the support of $\mathbf{h}$. Thus, we can replace $\mathbf{E}_p^{tr}$ with $\mathbf{E}^{te}$ and the equality still holds, we get: $\mathbf{T}(\mathbf{E}^{\text{te}}) = \mathbf{A}\mathbf{h}(\mathbf{f}(\mathbf{E}^{\text{te}})) + \mathbf{b} = \mathbf{A}\mathbf{h}(\mathbf{X}^{\text{te}}) + \mathbf{b}$. □

**Theorem 2**

*Proof.* After alignment, cause variable for each training pair is input to $\mathbf{h}$ as the first argument. By A2 and Theorem 1, we will successfully learn $\mathbf{h}$ (Algorithm 1, line 1,2).

By A2 and Corollary 1, if the cause variable of $\mathbf{X}^{te}$ is input to $\mathbf{hICA}$ as the first argument, then its nonlinear ICA is realized (Algorithm 1, line 3,4). Denote the respective input permutation as $\alpha_r$, then $C_{\alpha_r(1)} \perp\!\!\!\perp C_{\alpha_r(2)}$. While for the other input direction $\alpha_{1-r}$, by A1, $C_{\alpha_{1-r}(1)} \not\perp\!\!\!\perp C_{\alpha_{1-r}(2)}$

Thus, we have $\texttt{dindep}(\mathbf{C}_{\alpha_r}) > \texttt{dindep}(\mathbf{C}_{\alpha_{1-r}})$, and $\alpha^* = \alpha_r$. □

**Theorem 3**

*Proof.* Similarly to the proof of Theorem 2, we know there is one and only one input direction $\alpha_r$ where nonlinear ICA is realized. We have $\mathbf{T}(\mathbf{E}^{\text{te}}) = (C_{\alpha(1)}, C_{\alpha(2)})^T_{\alpha_r}$ where $\alpha$ is the unknown output permutation.

By A1 (which also implies rule (4)), we have $X_c^{te} \perp\!\!\!\perp C_{3-c,\alpha_r}$ where $c$ is the cause index, but $X_i^{te} \not\perp\!\!\!\perp C_{j,\alpha}$ for all other $i, j, \alpha$. Thus, $(i^*, j^*, \alpha^*) = (c, 3-c, \alpha_r)$ □

**Proposition 1**

*Proof.* From Definition 1, we write $\mathbf{X} = \mathbf{f}(\mathbf{E})$ and denote $\mathbf{g} = \mathbf{f}^{\text{-1}}$. And we have the relation of Jacobians $\mathbf{J_g} = \mathbf{J_f^{-1}}$, and:

$$\mathbf{J_f^{-1}} = \begin{pmatrix} \frac{df_1}{dE_1} & 0 \\ \frac{\partial f_2}{\partial X_1}\frac{\partial f_1}{\partial E_1} & \frac{\partial f_2}{\partial E_2} \end{pmatrix}^{-1}$$
$$= \begin{pmatrix} (\frac{df_1}{dE_1})^{-1} & 0 \\ -(\frac{\partial f_2}{\partial E_2})^{-1}\frac{\partial f_2}{\partial X_1}\frac{df_1}{dE_1}(\frac{df_1}{dE_1})^{-1} & (\frac{\partial f_2}{\partial E_2})^{-1} \end{pmatrix}$$

By comparing the 1st row of $\mathbf{J_g}$ and $\mathbf{J_f^{-1}}$, we have $\frac{\partial g_1}{\partial X_2} = 0$ which indicates $g_1$ is not a function of $X_2$, and $\frac{dg_1}{dX_1} = (\frac{df_1}{dE_1})^{-1}$ which, by inverse function theorem, implies $f_1$ is invertible and $g_1 = f_1^{-1}$. □

# 8 Nonlinear ICA violates causal faithfulness assumption

Causal Markov and faithfulness assumptions are common in causal discovery literature, and we also require them in our theorem. However, we should note that causal faithfulness assumption is violated for a realized bivariate nonlinear ICA, because the nonlinear ICA procedure necessarily has one of the following causal graphs:

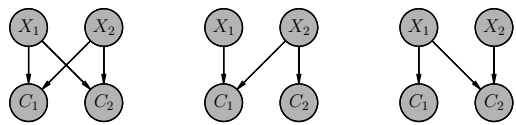

Figure 5: Causal graphs of nonlinear ICA procedure.

None of them induce $C_1 \perp\!\!\!\perp C_2$ under causal faithfulness assumption.

# 9 Choice of independence test

HSIC is a widely used independence test in causal discovery literature, but it has several drawbacks. First, its test statistic is not normalized for different testing pairs, and thus not comparable[9]. Second, although p-value of the test is comparable, it does not directly measure the degree of independence. Most importantly, as mentioned in Mooij et al. (2016, sec. 2.2), standard threshold of the test would be too tight for our purpose. This is because in causal discovery we often want to test the independence between an observed variable and an estimation *from* observed data, and there always exists small dependence with finite sample and other real world limitations. For the same reason, the flexibility of HSIC to detect dependence can do harm, not benefit, to causal discovery.

Unlike HSIC[10], dCor value is always in [0, 1], and equals to 0 if and only if the pair under test are independent. Thus, the value 1 - dCor works as a comparable degree of independence. As a bonus, dCor is much faster than HSIC when testing independence

---

[9] If we use the default Gaussian kernel and median heuristic for kernel bandwidth (Gretton et al., 2005). And this is also the most common way it is used in bivariate causal discovery (Mooij et al., 2016; Hu et al., 2018)

[10] We noticed that distance covariance is an instance of HSIC for certain choice of kernels (Sejdinovic et al., 2013). But again, this is not default for HSIC.

between univariate real-valued variables, particularly when sample size is large [11].

Hence, we suggest dCor rather than HSIC as the default choice to measure degree of independence for cause-effect inference, and try HSIC when you can afford the time, both for tuning and running.

## 10 Without assuming direct causal effect

We also experiment without assuming direct causal effect necessarily exists, and allow "inconclusive" outputs when the assumption is possibly violated. The purpose here is mainly to conform the problem mentioned in S.3 above, and to show how our method can address it to a large extent. When applying the inference rules, now we need to set a threshold or alpha value for the independence tests. For clearer comparisons, we apply Theorem 3 and also use HSIC, though Theorem 2 or other independence tests can also be applied. Then our method only differs with NonSENS by inferring for each environment and then majority voting.

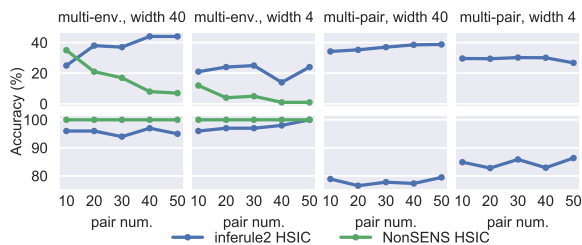

Figure 6: Performance without assuming direct causal effect. 1st/2nd row is results on direct casual data/purely confounded data respectively.

Similarly to Monti et al. (2019), we evaluate on two datasets: 1) all pairs are direct causal (1st row). 2) all pairs are purely confounded (simply use a fully connected MLP) (2nd row). On direct causal pairs, we can see NonSENS' accuracy decreases drastically w.r.t pair number and is nearly always below 10% when MLP width is 4. On the other hand, on purely confounded pairs, it always reports 100% inconclusive.

Here the results conform that the default alpha value (0.05) for independence test is way too tight. Specifically, the problem here is that, with more pairs (which means more sample points for NonSENS), HSIC is more sensitive to small dependence between estimated noise and observed cause. This means we must train

---

[11] We use Huo and Székely (2016) for dCor and Zhang et al. (2018) for HSIC, the implementation can be found at https://github.com/vnmabus/dcor and https://github.com/oxmlcs/kerpy, respectively.

TCL very optimally to avoid the unwanted dependence.

Our method performs much better than NonSENS, especially with large pair number. The reason is that, it is earlier to get rid of unwanted dependence by looking at each environment, since if any one of the environments shows dependence, then the pooled data tested in NonSENS will be dependent.

## 11 Additional notes on experiment on artificial data

**MLP width** The experimental results show that we need large enough MLP to fit more pairs. Note in particular that the MLP of width 4 performs almost always worse than that of width 40. If we use asymmetric MLP, this tendency is more drastic since it has much less parameters. When the MLP width is 4, the accuracy often decreases w.r.t the number of training pairs. When the MLP width is 40, the accuracy usually increases w.r.t the number of training pairs, but when the pair size is larger than 30, it increases slowly or even slightly drops.

**Training pair number** We observe better performance as the pair size grows (under the MLP width 40). Under the multi-pair setting, this implies that TCL learns more thoroughly the shared mechanism. Under multi-environment setting, we have one more reason: majority voting performs better with more voters (pairs).

## 12 Alternative ensemble scorings

Without loss of generality, assume $X_1$ is input to the same node when calculating $w_{ns,1}$, as cause variable is when training. Then we have $Direction_{ns} = \mathbb{I}(w_{ns,1} > w_{ns,2}) - \mathbb{I}(w_{ns,1} < w_{ns,2})$ where function $\mathbb{I}$ maps $true/false$ to $1/0$.

Now the ensemble score in Algorithm 4 line 10 becomes:

$$Score_s = \sum_{n \in TSR_s} w_{ns,1} w_n \mathbb{I}(w_{ns,1} > w_{ns,2}) \\ - \sum_{n \in TSR_s} w_{ns,2} w_n \mathbb{I}(w_{ns,1} < w_{ns,2}) \quad (5)$$

But since $\mathbb{I}(w_{ns,1} > w_{ns,2})$ and $\mathbb{I}(w_{ns,1} < w_{ns,2})$ just reflect the relative value of $w_{ns,1}$ and $w_{ns,2}$, the following simplification is reasonable:

$$Score_s = \sum_{n \in TSR_s} w_n (w_{ns,1} - w_{ns,2}) \quad (6)$$

And on the same line of reasoning, we can disregard

$w_{ns,1}, w_{ns,2}$ and have:

$$Score_s = \sum_{n \in TSR_s} w_n Direction_{ns} \quad (7)$$

This is just the weighted average of prediction by each $\mathbf{h}_n$. And finally, since $\mathbf{h}_n$ with small $w_n$ is unlikely to produce large $w_{ns,i}$, we can further disregard $w_n$ in (6). This gives:

$$Score_s = \sum_{n \in TSR_s} (w_{ns,1} - w_{ns,2}) \quad (8)$$

We compared these scoring equations and found (4) is stably the best.